\documentclass{article}

\usepackage{arxiv}

\usepackage[utf8]{inputenc}
\usepackage[T1]{fontenc}    
\usepackage{hyperref}      
\usepackage{url}
\usepackage{booktabs}       
\usepackage{amsfonts}       
\usepackage{microtype}
\usepackage{fancyhdr}  
\usepackage{graphicx}
\usepackage{latexsym}
\usepackage{booktabs}
\usepackage{enumitem}
\usepackage{color}
\usepackage{amsmath}
\usepackage{amssymb}
\usepackage{amsthm}
\usepackage{import}
\usepackage{float}
\usepackage{tabularx}
\usepackage{caption}
\usepackage{subcaption}
\usepackage{algorithm}
\usepackage{algpseudocode}

\title{Concept Boundary Vectors}

\author{
  Thomas Walker\\
  Independent\\
  \texttt{thomas.mattia.walker@hotmail.co.uk}
}

\begin{document}
\maketitle

\begin{abstract}
    Machine learning models are trained with relatively simple objectives, such as next token prediction. However, on deployment, they appear to capture a more fundamental representation of their input data. It is of interest to understand the nature of these representations to help interpret the model's outputs and to identify ways to improve the salience of these representations. Concept vectors are constructions aimed at attributing concepts in the input data to directions, represented by vectors, in the model's latent space. In this work, we introduce concept boundary vectors as a concept vector construction derived from the boundary between the latent representations of concepts. Empirically we demonstrate that concept boundary vectors capture a concept's semantic meaning, and we compare their effectiveness against concept activation vectors.
\end{abstract}

\section{Introduction}

Machine learning models are commmonly trained using comparatively basic objectives. For instance, large language models are optimised to `merely' predict the next token in a sequence of text \cite{vaswani_attention_2023}. Despite this simple objective, the model can generate syntactically and semantically coherent text \cite{radford_language_2018}. Similarly, word embedding models, such as Word2Vec \cite{mikolov_efficient_2013}, are trained to minimise the distance between contextually similar words. By investigating the resulting model, we observe that non-trivial relationships between words are coherently captured in the embedding space \cite{mikolov_efficient_2013}.
Therefore, since there is an increasing application of machine learning models in the life sciences \cite{jumper_highly_2021,tandon_systematic_2024}, safety-sensitive domains \cite{kouvaros_formal_2021} and the general population \cite{openai_introducing_2022}, it is important to be able to interpret the model's internal processing. Numerous \emph{interpretability methods} have been designed to~(partially) uncover the internal workings of a machine learning model \cite{ribeiro_why_2016,lundberg_unified_2017,kim_interpretability_2018}. More specifically, inner interpretability methods focus on the internal structures and representations to explain model behaviour \cite{rauker_toward_2023}. 

In this work, we explore how concepts are represented in the latent spaces of models. These concepts are either pre-determined concepts that are thought to be pertinent to the functioning of the model, or they are determined in an unsupervised manner. Methods for the latter include sparse autoencoders, which have been shown to successfully extract interpretable concepts from the latent spaces of large language models \cite{cunningham_sparse_2023}. Here, we explore \emph{concept vectors}, which are supervised constructions of vectors that attempt to encode the semantic meaning of a pre-defined concept in the internal representations of a model.

\subsection{Related Work}

Concept activation vectors \cite{kim_interpretability_2018} is a concept vector construction utilising a linear classifier trained to distinguish the latent representations of data with and without the concept present. Concept activation vectors have been shown to effectively represent concepts in vision models \cite{kim_interpretability_2018}, and have been used to analyse deployed machine learning models \cite{lucieri_interpretability_2020}.  

Adversarial concept activation vectors (A-CAV) and orthogonal adversarial concept activation vectors (OA-CAV) \cite{soni_adversarial_2020} are an extension of concept activation vectors aimed at improving the effectiveness and robustness of the concept vector.

Concept activation vectors are constructed under the assumption that the concept and non-concept latent representations can be linearly separated. Concept activation regions \cite{crabbe_concept_2022} are derived to account for the possibility that this assumption does not hold. To increase the validity of this assumption, we only consider relative concept vectors which are constructed from sets of data corresponding to distinct, distinguished concepts. Therefore, our concept vectors will represent the relationship between concepts.

Our work is most similar to concept activation regions \cite{crabbe_concept_2022}, in that we consider the geometry of the latent representations of concepts to inform the construction of the concept vector. 

\subsection{Contributions}

We introduce concept boundary vectors as an alternative concept vector formalism, that is motivated to be faithful to the geometry of the boundary between concepts. We introduce this construction in an attempt to better capture the relationship between concepts, since existing work shows that for classification models, the geometrical complexity of a decision boundary influences the model's accuracy \cite{ramamurthy_topological_2018}. We hypothesise that much of the semantic meaning of a concept-concept relationship is encoded in the geometry of the boundary between the latent representations of these concepts.


\section{Concept Vectors}\label{sec:concept_vectors}

Concept vectors are constructions that attempt to represent the relationship of a concept relative to other concepts in the latent space of a machine learning model \cite{rauker_toward_2023}. They should be thought of as \emph{directions} in the latent space of a model, pointing in a direction that is positively associated with a concept. Concept vectors are typically constructed in a supervised manner, requiring the availability of data where the concept is present, and data where the concept is not present.
The expectation that concept vectors can faithfully represent the latent space activations of a concept is predicated on some assumptions:
\begin{enumerate}
    \item[A1.] The latent space activations of different concepts are linearly separable.
    \item[A2.] The collective latent space activations of a concept exhibit some homogeneity. That is, a single vector exists that can capture the majority of the semantic meaning of a concept.
\end{enumerate}

A1 is usually formulated as the \textit{linear representation hypothesis} \cite{park_linear_2024}, and has empirically developed after subsequent pieces of work have successfully utilised it to extract meaningful features from machine learning models \cite{mikolov_efficient_2013,cunningham_sparse_2023,nanda_emergent_2023,hernandez_inspecting_2023,yang_local_2023}.

A2 on the other hand is less frequently explored; we rectify this using topological data analysis, Section \ref{sec:concept_homogeneity}. 

\subsection{Concept Activation Vectors}

Concept activation vectors \cite{kim_interpretability_2018} are a concept vector construction that depends significantly on the validity of A1. Let $f:\mathbb{R}^d\to\mathbb{R}^k$ represent a neural network. Let $f_{\ell}:\mathbb{R}^d\to\mathbb{R}^m$ be the function of the network from the input to the $\ell^{\text{th}}$ layer output. For a concept $\mathcal{C}$ we can construct a set of inputs $S_{\mathcal{C}}\subseteq\mathbb{R}^d$ that contain said concept. From $S_{\mathcal{C}}$ and $f_{\ell}$ we can obtain the latent representation of the concept $\mathcal{C}$ at layer $\ell$ of the model as $$A_{\mathcal{C}}=\left\{f_{\ell}(s):s\in S_{\mathcal{C}}\right\}.$$For concepts $\mathcal{C}_{+}$ and $\mathcal{C}_{-}$,\footnote{We will interchangeably refer to $\mathcal{C}_+$ as the positive or target concept, and $\mathcal{C}_-$ as the negative or source concept.} using $A_{\mathcal{C}_{+}}$ and $A_{\mathcal{C}_{-}}$ we train a linear classifier\footnote{Using the binary cross entropy loss.} $v_{\pm}\in\mathbb{R}^m$ to satisfy $$\begin{cases}\sigma\left(v_{\pm}\cdot a_+\right)\geq0.5&a_+\in A_{\mathcal{C}_{+}}\\\sigma\left(v_{\pm}\cdot a_-\right)<0.5&a_-\in A_{\mathcal{C}_{-}},\end{cases}$$where $\sigma(\cdot)$ is the sigmoid function. The vector $v_{\pm}$ is the (relative) concept activation vector from concept $\mathcal{C}_{-}$ to concept $\mathcal{C}_{+}$. We assume this terminology since geometrically $v_{\pm}$ can be thought of as a vector pointing in the direction of the latent representations of concept $\mathcal{C}_+$ from the latent representations of the concept $\mathcal{C}_-$.

\subsection{Concept Boundary Vectors}

As before, suppose we have concepts $\mathcal{C}_+$ and $\mathcal{C}_-$ with corresponding latent representations $A_{\mathcal{C}_+}$ and $A_{\mathcal{C}_-}$. We employ Algorithm \ref{alg:boundary_construction} on $A_{\mathcal{C}_+}$ and $A_{\mathcal{C}_-}$ to identify pairs $\left(a_+,a_-\right)\in P_{\pm}\subseteq A_{\mathcal{C}_+}\times A_{\mathcal{C}_{-}}$ that mark the \emph{boundary} between the positive concept and the negative concept. Using these tuples we can construct \textit{boundary normal vectors}, $$\mathcal{N}_{\pm}:=\left\{\frac{a_+-a_-}{\left\Vert a_+-a_-\right\Vert_2}:(a_+,a_-)\in P_{\pm}\right\}.$$We then optimize a vector to be similar, in terms of cosine similarities, to the vectors of $\mathcal{N}_{\pm}$.\footnote{The loss function for this optimization is the absolute value of one minus the dot product of this vector with the boundary normal vectors.} We term the resulting vector the concept boundary vector for the relationship of $\mathcal{C}_-$ to $\mathcal{C}_+$.

\subsection{Properties}

\subsubsection{Optimization}

\begin{figure}[ht]
     \centering
     \begin{subfigure}[b]{0.38\textwidth}
         \centering
         \includegraphics[width=0.65\textwidth]{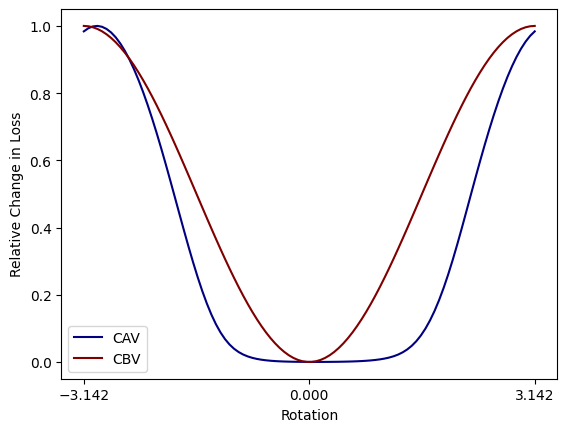}
         \caption{}
         \label{fig:2d_concept_vector_loss_variability}
     \end{subfigure}
     \hfill
     \begin{subfigure}[b]{0.58\textwidth}
         \centering
         \includegraphics[width=0.75\textwidth]{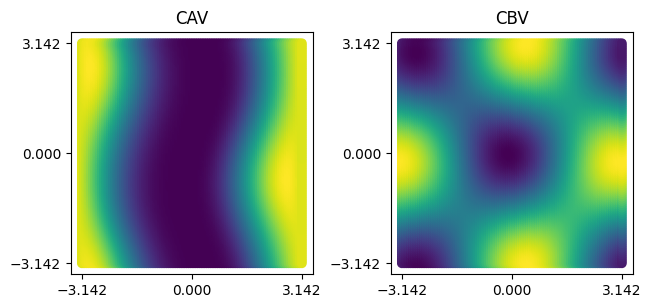}
         \caption{}
         \label{fig:3d_concept_vector_loss_variability}
     \end{subfigure}
        \caption{Figure \ref{fig:2d_concept_vector_loss_variability} shows the variability in the loss as a concept vector trained two-dimensional data is rotated. Figure \ref{fig:3d_concept_vector_loss_variability} shows the variability in the loss as a concept vector trained on three-dimensional data is rotated.}
        \label{fig:concept_vector_loss_variability}
\end{figure}

The optimisation objective of maximising similarity -- imposed to obtain the concept boundary vector -- is often more restrictive than the classification objective -- imposed to obtain the concept activation vector. This is highlighted in Figure \ref{fig:concept_vector_loss_variability} which visualises the respective losses for concept vectors as they are rotated in space. In the two-dimensional case, the similarity loss increases more sharply, whereas the classification loss remains relatively low for a non-trivial amount of rotation. In three dimensions this effect is only magnified, Figure \ref{fig:3d_concept_vector_loss_variability}.

\subsubsection{Computational Complexity}

Algorithm \ref{alg:boundary_construction} provides an additional computational burden to computing a concept vector. We explore the extent of this burden in Section \ref{sec:boundary_construction_algorithm}. However, we will demonstrate that this can be justified as concept boundary vectors provide a more faithful representation of the relationship between concepts.

\subsubsection{Logit Influence}

From our intuition that concept vectors identify the relationship between concepts, perturbing activations in these directions should predictably influence the outputted logits of the model. For a classification model, perturbing the activations with the concept vector corresponding to classes of the model should lead to the logit of the positive class increasing and the logit of the negative class decreasing. We quantify this with the logit influence (LI) metric, which is just an extension of the TCAV score \cite{kim_interpretability_2018}.

Suppose $v^{\ell}_{\pm}$ is a concept vector extracted from the representations of concepts $\mathcal{C}_+$ and $\mathcal{C}_-$ at layer $\ell$ of a model. Then the logit influence of $v^{\ell}_{\pm}$ on class $\mathcal{C}$ is 
\begin{equation}\label{eq:logit_influence}
    \mathrm{LI}_{\mathcal{C}}\left(v_{\pm}^{\ell}\right)=\frac{1}{\left\vert A_{\mathcal{C}_+}\right\vert}\sum_{a_+\in A_{\mathcal{C}_+}}\lim_{\epsilon\to0}\left(\frac{f_{\ell,\mathcal{C}}\left(a_++\epsilon v^{\ell}_{\pm}\right)-f_{\ell,\mathcal{C}}\left(a_+\right)}{\epsilon}\right),
\end{equation}
where $f_{\ell,\mathcal{C}}$ denotes the function of the model from layer $\ell$ to the output logit corresponding to class $\mathcal{C}$. In other words, the logit influence metric calculates the average gradient of the model from layer $\ell$ to the output logit corresponding to class $\mathcal{C}$ in the direction of $v_{\pm}^{\ell}$.

\section{Concept Vector Analysis}\label{sec:experiments}

With our construction, we now explore concept vectors obtained from a convolutional neural network trained on MNIST \cite{deng_mnist_2012}. Understanding the evolution of concept vectors through the layers of a model is also interesting, but requires a larger model trained on a sufficiently complex task. Hence, for exposition, we focus on a simpler network and only extract the concept vectors at a single hidden layer. In Section \ref{sec:vit} we explore concept vectors from a vision transformer model \cite{dosovitskiy_image_2021} trained on CIFAR10 \cite{krizhevsky_learning_2009}.\footnote{The code accompanying these investigations is available \hyperlink{https://github.com/ThomasWalker1/concept_boundary_vectors/tree/main}{here}.}

\subsection{Experimental Details}

Our convolutional neural network consists of two convolutional layers, a max pooling layer, a linear embedding layer and then a linear output layer. Between each layer, there is a ReLU non-linearity. The network is trained on the MNIST dataset \cite{deng_mnist_2012} for 5 epochs with a batch size of 256 and a learning rate of 0.001. The Adam optimizer is used with the default values provided by PyTorch \cite{pytorch_contributors_adam_2023}. We consider each digit as a concept and extract the 64-dimensional latent representations for these concepts after the linear embedding layer of the network. For each concept, $S_{\mathcal{C}}$ contains the inputs that are correctly classified by the trained model. 

For concepts $\mathcal{C}_+$ and $\mathcal{C}_-$, the concept activation vector is obtained by training a 64-dimensional linear classifier. We use a batch size of 64, a learning rate of 0.001 and 100 epochs to train the classifier to distinguish the sets $A_{\mathcal{C}_+}$ and $A_{\mathcal{C}_-}$.

For concepts $\mathcal{C}_+$ and $\mathcal{C}_-$, the concept boundary vector is obtained by optimizing a vector to maximize the average similarity to the set of boundary normal vectors $\mathcal{N}_{\pm}$. We optimize at individual instances of $\mathcal{N}_{\pm}$ a total of 10000 times.

For the moment we will assume the validity of A1 and A2, and explore the properties of the resulting concept vectors. Then in Section \ref{sec:top_analysis_concept_vetors} we will investigate these assumptions using topological data analysis.

\subsection{Logit Influence}

For each concept-concept relationship, we compute the logit influence of the concept activation vector and the concept boundary vector on the target and source class.\footnote{Before this analysis, we check that each concept vector influences its target class more significantly than the other classes.}

\begin{figure}[ht]
     \centering
     \begin{subfigure}[b]{0.48\textwidth}
         \centering
         \includegraphics[width=0.5\textwidth]{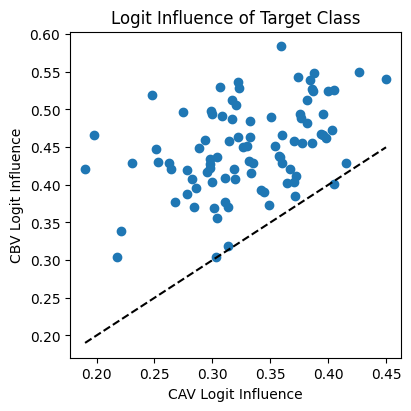}
         \caption{}
         \label{fig:logit_influence_on_target}
     \end{subfigure}
     \hfill
     \begin{subfigure}[b]{0.48\textwidth}
         \centering
         \includegraphics[width=0.5\textwidth]{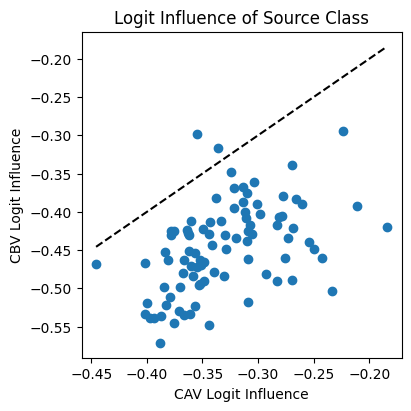}
         \caption{}
         \label{fig:logit_influence_on_source}
     \end{subfigure}
        \caption{Figure \ref{fig:logit_influence_on_target} shows the influence of the concept vectors on the logit of the target concept. Figure \ref{fig:logit_influence_on_source} shows the influence of the concept vectors on the logit of the source concept. The dashed line is the line where the influence of the vectors is equal.}
        \label{fig:logit_influence}
\end{figure}

From Figure \ref{fig:logit_influence_on_target} we observe that in almost all cases the concept boundary vector has a greater logit influence on the target class. Suggesting that the concept boundary vector is a more effective representation of the concept-concept relationship. This is further evidenced by Figure \ref{fig:logit_influence_on_source} which shows that the concept boundary vector more significantly suppressed the source concept.

\subsection{Concept Entanglement}

We should expect that similar concepts are to some extent \emph{entangled} in the latent representations of a model. Recall that, when constructed between well-defined concepts as we are doing here, the concept vector encodes the relationship from the negative concept to the positive concept. Therefore, as we are considering concepts to be the different digits of MNIST, a concept vector can be thought of as encoding how to transform the negative digit into the positive digit. Hence, by similar concepts we mean the transformation from the negative to the positive digit is similar. 

We first study concept entanglement from a Euclidean perspective, that is using cosine similarity as a measure of similarity. However, we then take a more topological perspective due to the evidence that concept vectors do not necessarily naturally live in Euclidean space \cite{park_linear_2024}. More specifically, we extend the analysis by considering concept vectors as points on a high-dimensional unit sphere and studying the topological properties of this representation.

\subsubsection{Euclidean Perspective}

\begin{figure}[ht]
    \centering
    \includegraphics[width=0.5\linewidth]{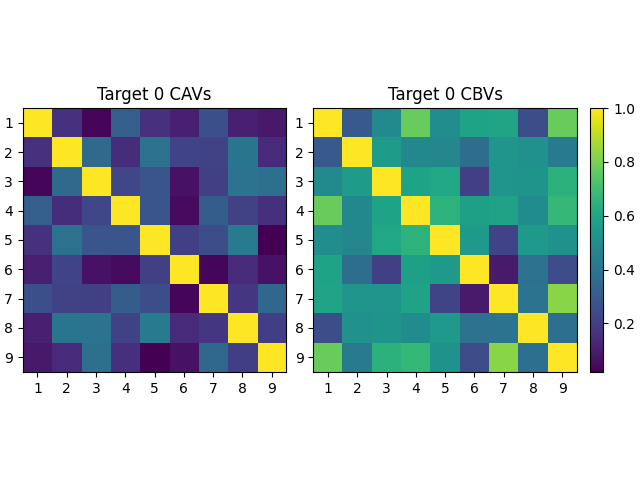}
    \caption{The cosine similarities between concept vectors with the concept $\mathsf{0}$ as their target.}
    \label{fig:entanglement_target}
\end{figure}

In Figure \ref{fig:entanglement_target} we depict the cosine similarities between concept vectors that have the target concept $\mathsf{0}$. Immediately we observe that the cosine similarities between concept boundary vectors are more varied. In particular, the relationship to the concept $\mathsf{6}$ is dissimilar to the relationship to the concept $\mathsf{7}$; perhaps reflective of the fact that the process of transforming a $\mathsf{6}$ to a $\mathsf{0}$ is very different to the process of transforming a $\mathsf{7}$ to a $\mathsf{0}$. On the other hand, the relationship to the concept $\mathsf{9}$ is similar to the relationship to the concept $\mathsf{7}$; perhaps because the process of transforming these concepts into a $\mathsf{0}$ is similar. Concept boundary vectors seem to be capturing more of the semantic meaning present in these concept-concept relationships.

\subsubsection{Persistent Homology}

To construct a holistic understanding of the collection of concept vectors we apply persistent homology to the point cloud on the unit sphere using the geodesic distance.

\begin{figure}[ht]
    \centering
    \includegraphics[width=0.5\linewidth]{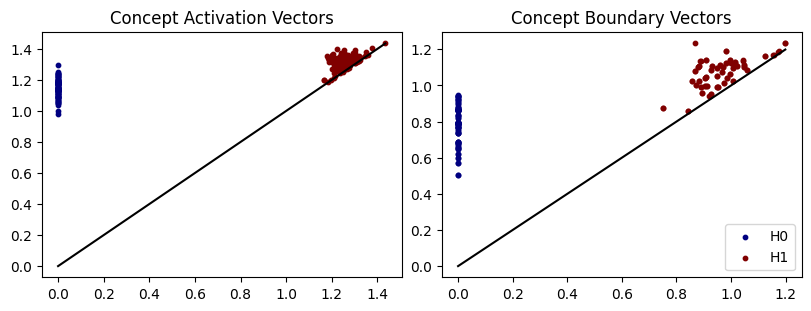}
    \caption{The persistence diagrams obtained from geodesic-based filtrations of concept activation vectors and concept boundary vectors.}
    \label{fig:persistence_diagram}
\end{figure}

From Figure \ref{fig:persistence_diagram} it is evident that the topological features of concept boundary vectors are more distinct than those of concept activation vectors. The one-dimensional features, which correspond to loop-style structures, are more persistent in the case of concept boundary vectors. This chain-like structure is perhaps to be expected since concept vectors with the same target digit are likely to be entangled. When concept vectors with different targets are also entangled, these chains of concept vectors connect; resulting in structures that are likely to persist through the one-dimensional homology groups.

\subsubsection{The Mapper Algorithm}

To obtain conclusions regarding the interactions between individual concept vectors, we use the mapper algorithm \cite{singh_topological_2007} to construct a graphical low-dimensional representation of the concept vectors which preserves some of its underlying topology.\footnote{For the experiments we use Kepler Mapper \cite{hendrik_jacob_van_veen_kepler_2019}, which is a Python implementation of the mapper algorithm.}


\begin{figure}[ht]
    \centering
    \includegraphics[width=0.75\linewidth]{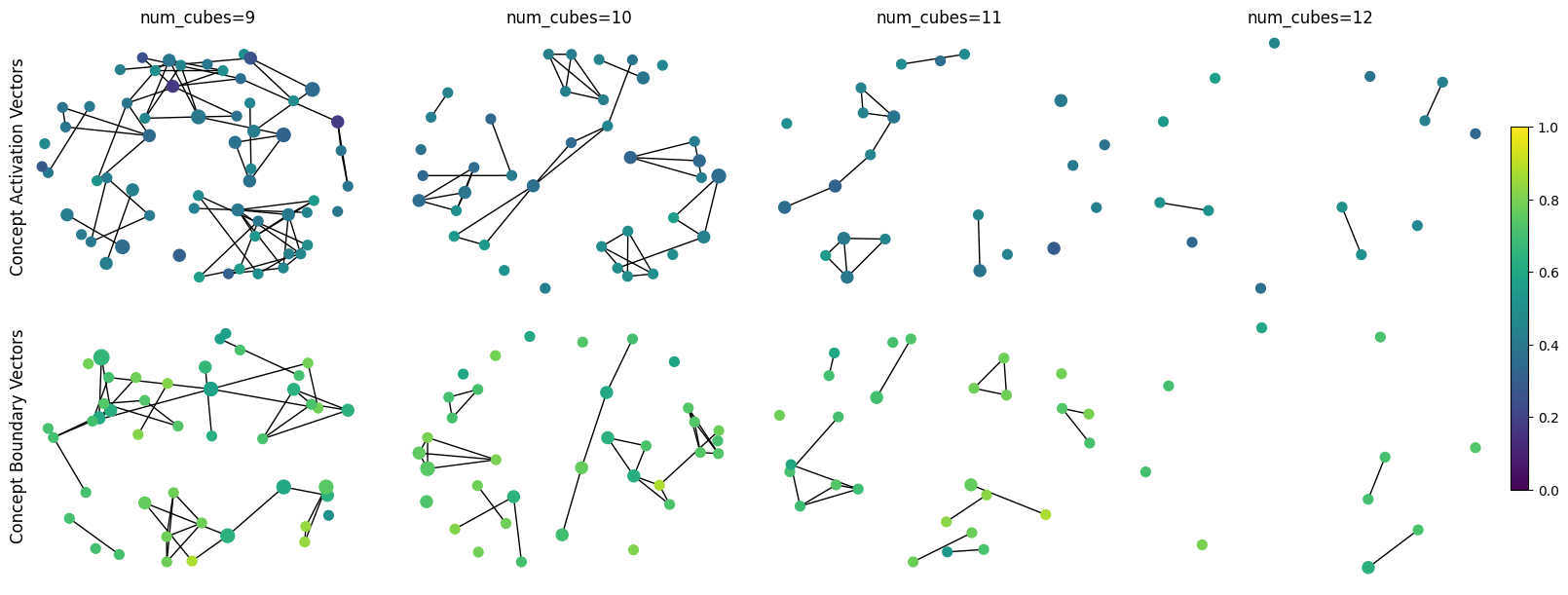}
    \caption{Mapper plots obtained from concept activation vectors and concept boundary vectors. The size of the dots represents the number of concept vectors in the cluster represented by the node. The colour of the nodes is the average cosine similarity between the concept nodes within the cluster represented by the node.}
    \label{fig:mapper_graphs}
\end{figure}

Figure \ref{fig:mapper_graphs} shows a topological structure within concept activation and boundary vectors. It is perhaps more distinct in the case of concept boundary vectors as it arises for a lower number of cubes, meaning we can extract features at a lower resolution. Moreover, for concept boundary vectors we see that the vectors within the cluster have a higher cosine similarity.

In Table \ref{tab:mapper_graph_clusters} we analyse in more detail the graphs obtained with a cover consisting of 11 cubes.

\begin{table}[ht]
    \centering
    \begin{tabular}{|p{3cm}|p{5cm}|p{6cm}|}
        \hline
        Concept Vector& Clusters & Simplices \\
        \hline
        \hline
        Activation Vectors & $(\mathsf{8\_3},\mathsf{8\_5}),\,(\mathsf{1\_0},\mathsf{1\_6}),\,(\mathsf{1\_4},\mathsf{3\_4})$, $(\mathsf{5\_2},\mathsf{5\_3}),\,(\mathsf{7\_0},\mathsf{7\_6}),\,(\mathsf{1\_7},\mathsf{1\_9})$, $(\mathsf{4\_5},\mathsf{9\_5}),\,(\mathsf{8\_0},\mathsf{8\_6}),\,(\mathsf{4\_0},\mathsf{4\_6})$, $(\mathsf{2\_5},\mathsf{2\_6}),\,(\mathsf{6\_2},\mathsf{6\_7}),\,(\mathsf{2\_4},\mathsf{2\_9})$, $(\mathsf{1\_3},\mathsf{1\_5},\mathsf{1\_8}),\,(\mathsf{3\_6},\mathsf{5\_6},\mathsf{9\_4})$, $(\mathsf{4\_8},\mathsf{9\_0},\mathsf{9\_8}),\,(\mathsf{4\_5},\mathsf{7\_5},\mathsf{9\_5})$, $(\mathsf{4\_0},\mathsf{9\_0},\mathsf{9\_8}),\,(\mathsf{4\_6},\mathsf{7\_0},\mathsf{7\_6})$, $(\mathsf{8\_3},\mathsf{8\_5},\mathsf{8\_6})$ & $(\mathsf{4\_5},\mathsf{7\_5},\mathsf{9\_5})-(\mathsf{4\_5},\mathsf{9\_5})$,\newline$(\mathsf{4\_0},\mathsf{4\_6})-(\mathsf{4\_6},\mathsf{7\_0},\mathsf{7\_6})-(\mathsf{7\_0},\mathsf{7\_6})-(\mathsf{4\_0},\mathsf{9\_0},\mathsf{9\_8})-(\mathsf{4\_8},\mathsf{9\_0},\mathsf{9\_8})$,\newline$(\mathsf{8\_0},\mathsf{8\_6})-(\mathsf{8\_3},\mathsf{8\_5},\mathsf{8\_6})-(\mathsf{8\_3},\mathsf{8\_5})$\\
        \hline
        Boundary Vectors & $(\mathsf{1\_5},\mathsf{1\_6}),\,(\mathsf{8\_3},\mathsf{8\_5}),\,(\mathsf{0\_1},\mathsf{0\_4})$, $(\mathsf{5\_4},\mathsf{8\_4}),\,(\mathsf{4\_1},\mathsf{7\_1}),\,(\mathsf{9\_0},\mathsf{9\_8})$, $(\mathsf{9\_1},\mathsf{9\_7}),\,(\mathsf{3\_6},\mathsf{5\_6}),\,(\mathsf{2\_3},\mathsf{7\_3})$, $(\mathsf{3\_1},\mathsf{7\_1}),\,(\mathsf{3\_7},\mathsf{8\_7}),\,(\mathsf{3\_6},\mathsf{8\_6})$, $(\mathsf{7\_6},\mathsf{9\_6}),\,(\mathsf{9\_0},\mathsf{9\_6}),\,(\mathsf{5\_1},\mathsf{6\_1})$, $(\mathsf{4\_3},\mathsf{4\_5}),\,(\mathsf{0\_7},\mathsf{0\_9}),\,(\mathsf{5\_0},\mathsf{5\_6})$, $(\mathsf{2\_3},\mathsf{2\_5}),\,(\mathsf{4\_3},\mathsf{4\_5},\mathsf{9\_3})$, $(\mathsf{5\_4},\mathsf{5\_7},\mathsf{5\_9})$ & $(\mathsf{5\_4},\mathsf{5\_7},\mathsf{5\_9})-(\mathsf{5\_4},\mathsf{8\_4})$,\newline$(\mathsf{3\_6},\mathsf{5\_6})-(\mathsf{5\_0},\mathsf{5\_6})-(\mathsf{3\_6},\mathsf{8\_6})$,\newline$(\mathsf{9\_0},\mathsf{9\_8})-(\mathsf{9\_0},\mathsf{9\_6})-(\mathsf{7\_6},\mathsf{9\_6})$,\newline$(\mathsf{4\_3},\mathsf{4\_5})-(\mathsf{4\_3},\mathsf{4\_5},\mathsf{9\_3})$,\newline$(\mathsf{2\_3},\mathsf{7\_3})-(\mathsf{2\_3},\mathsf{2\_5})$\\ 
        \hline
    \end{tabular}
    \caption{For each concept vector formalism we identify the concept vectors contained within the clusters, represented by nodes, in the mapper graph with 11 cubes in its cover. Furthermore, we identify the clusters that form the simplices of the graph. The notation $\mathsf{0\_1}$ represent the concept vector whose target is $\mathsf{0}$ and whose source is $\mathsf{1}$. Some of the nodes contain the same concept vectors, we remove these duplicates in the presented data and that is why there are fewer identified clusters than nodes in the graph, and why some of the simplices have fewer identified clusters.}
    \label{tab:mapper_graph_clusters}
\end{table}

From Table \ref{tab:mapper_graph_clusters} we see that clusters for concept activation vectors and concept boundary vectors conform to our expectation that similar concept vectors are those where the transformation from the source digit to the target digit is similar. One could qualitatively argue that the clusters for concept boundary vectors are more coherent.

\subsection{Concept Algebra}

There is evidence that models encode concepts in ways that satisfy \emph{linear algebra} \cite{mikolov_efficient_2013}, in the sense that linear combinations of concepts give reasonable results. We would like to test this hypothesis for our model using concept vectors. Specifically, if we have concepts $\mathcal{C}_1$, $\mathcal{C}_2$ and $\mathcal{C}_3$ with concept vectors $v_{1,2}$ and $v_{2,3}$ as expected, then how similar is $v_{1,2}+v_{2,3}$ to $v_{1,3}$? 

If these concept vectors satisfy linear algebraic properties then we would expect this similarity to be high. Empirically we test this by computing this linear combination for a pair of concept vectors $v_{1,2}+v_{2,3}$ and comparing its similarity to concept vectors of the form $v_{1,\mathcal{C}}$, where $\mathcal{C}$ is some concept not equal to $\mathcal{C}_1$ or $\mathcal{C}_2$. If out of these vectors, the one most similar to $v_{1,2}+v_{2,3}$ is $v_{1,3}$ then we say that the concept algebra has been successful. More formally, concepts $\mathcal{C}_1$, $\mathcal{C}_2$ and $\mathcal{C}_3$ successfully exhibit linear algebraic properties if $$\mathcal{C}_3=\mathrm{argmax}_{\mathcal{C}}\left(\left(v_{1,2}+v_{2,3}\right)\cdot v_{1,\mathcal{C}}\right).$$

\begin{figure}[ht]
     \centering
     \begin{subfigure}[b]{0.48\textwidth}
        \centering
        \includegraphics[width=0.75\linewidth]{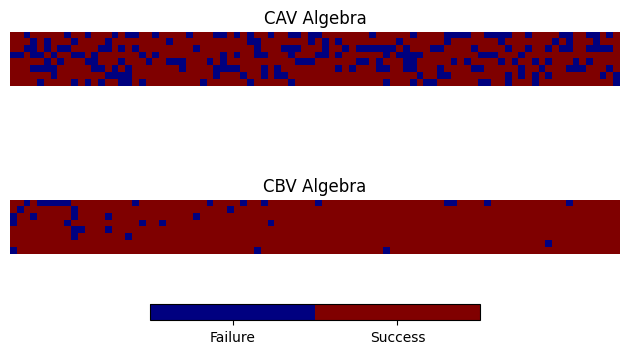}
        \caption{}
        \label{fig:concept_algebra_succes_rate}
     \end{subfigure}
     \hfill
     \begin{subfigure}[b]{0.48\textwidth}
        \centering
        \includegraphics[width=0.75\linewidth]{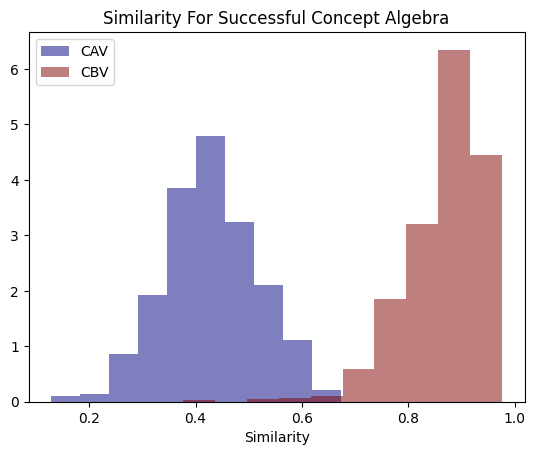}
        \caption{}
        \label{fig:concept_algebra_similarities}
     \end{subfigure}
        \caption{In Figure \ref{fig:concept_algebra_succes_rate} each column of the image plots corresponds to a concept-concept relationship. Each square in this column then represents whether the concept algebra involving one of the eight remaining concepts is a success. Figure \ref{fig:concept_algebra_similarities} shows the similarities between the combined concept vectors and the target concept vector when the concept algebra was successful.}
        \label{fig:concept_algebra}
\end{figure}

From Figure \ref{fig:concept_algebra_succes_rate} we see that in most cases the algebra of concept vectors is successful. In particular, the success rate is much higher for concept boundary vectors than for concept activation vectors. Providing evidence that concept boundary vectors are more coherent in their representations. Moreover, from Figure \ref{fig:concept_algebra_similarities} we see that the success is stronger in the case of concept boundary vectors, as the combined vectors are more similar to the target concept vector.

\subsection{Adversarial Construction}\label{sec:adversarial_construction}

Ultimately, the extent to which concept vectors effectively encode the relationship between concepts is determined by the extent to which they influence the model from which they are obtained. Given concepts $\mathcal{C}_+$ and $\mathcal{C}_-$ with concept vector $v_{\pm}$, we should be able to perturb an element of $A_{\mathcal{C}_-}$ with $v_{\pm}$ such that it is reclassified by the model as concept $\mathcal{C}_+$.

Here we consider latent representations $a_-\in A_{\mathcal{C}_-}$ that are components in a pair from $P_{\pm}$. We perturb the latent representations with a concept vector $v_{\pm}$ up to the point of reclassification. We then translate this perturbation into the input space by optimising for input noise added to the input corresponding to $a_-$ which would cause the necessary perturbation in the latent space.

\begin{figure}[ht]
    \centering
    \includegraphics[width=0.5\linewidth]{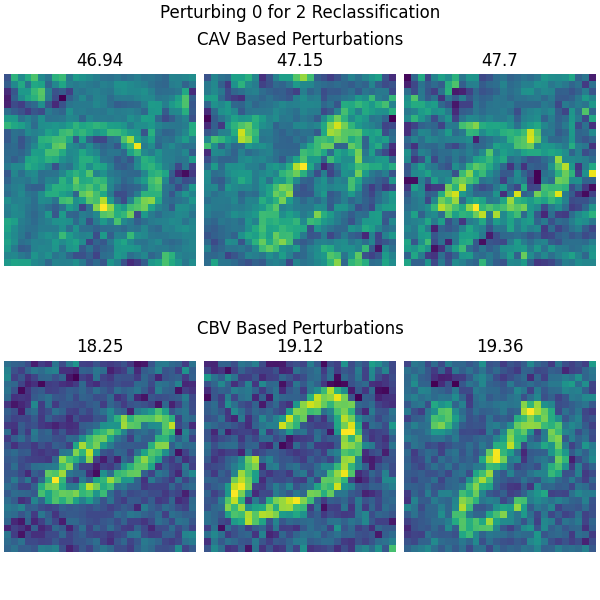}
    \caption{Perturbed inputs that are classified by $\mathsf{2}$ by the model despite originally representing the digit $\mathsf{0}$. The values on top of each input are the distances of the perturbed inputs from their original form.}
    \label{fig:image_perturbation}
\end{figure}

From Figure \ref{fig:image_perturbation} we see that concept boundary vector-based perturbations are more effective at reclassifying the latent representations of a concept since the corresponding perturbed inputs are closer to the original inputs.



It is important to note that the above perturbations are performed on latent vectors that are identified to be on the boundary by Algorithm \ref{alg:boundary_construction}. We do so to minimize the perturbation necessary to get reclassification, such that the resulting input still largely resembles the original input. In this sense, we have constructed adversarial inputs for the classifier using concept vectors. However, this raises the question as to whether the lower amplitude perturbation is just because the concept boundary vector overfit to the relatively few identified boundary points.

\begin{figure}[ht]
    \centering
    \includegraphics[width=0.3\linewidth]{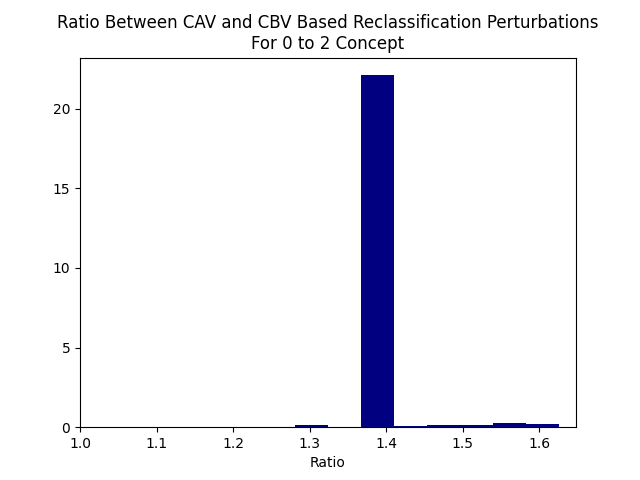}
    \caption{The ratio of the magnitude of concept activation vector-based perturbations and concept boundary vector-based perturbations required to reclassify a latent vector of concept $\mathsf{0}$ as a latent vector of concept $\mathsf{2}$.}
    \label{fig:ratio_perturbations}
\end{figure}

From Figure \ref{fig:ratio_perturbations} we observe that the effectiveness of concept boundary vector-based perturbations observed on the boundary latent activations extends to the entire cluster of latent activations. This is evidence that the concept boundary vector is not overfitting to the boundary latent activations, but instead encapsulates information that generalises across the entire cluster of latent activations. This generalisation would only be feasible if A2 is satisfied.

\subsection{Identifying Spatial Features}

Here, we would like to understand how the encoded semantics of our concept vector translate into the input space. The structure of the noise added to the inputs from concept vector-based perturbations in the latent space is not clear from Figure \ref{fig:image_perturbation}. We are motivated to conduct this investigation as previous work shows that concept vectors can meaningfully capture spatial information from the inputs they were obtained from \cite{nicolson_explaining_2024}. For each input corresponding to a negative activation in a pair of $P_{\pm}$, we optimize for an input perturbation of unit norm that induces the most significant perturbation in the latent space aligned with the concept-boundary vector. That is, for $s\in S_{\mathcal{C}_-}$ we optimize for an input perturbation $\delta s$ such that $\Vert\delta s\Vert_2=1$ and $f_{\ell}(s+\delta s)\cdot v^{\ell}_{\pm}$ is maximized.


\begin{figure}[ht]
     \centering
     \begin{subfigure}[b]{0.45\textwidth}
        \centering
        \includegraphics[width=\linewidth]{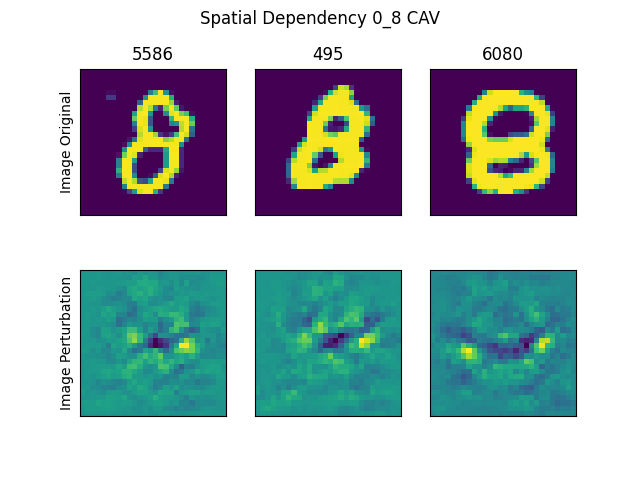}
        \caption{}
        \label{fig:spatial_dependency_cav}
     \end{subfigure}
     \hfill
     \begin{subfigure}[b]{0.45\textwidth}
        \centering
        \includegraphics[width=\linewidth]{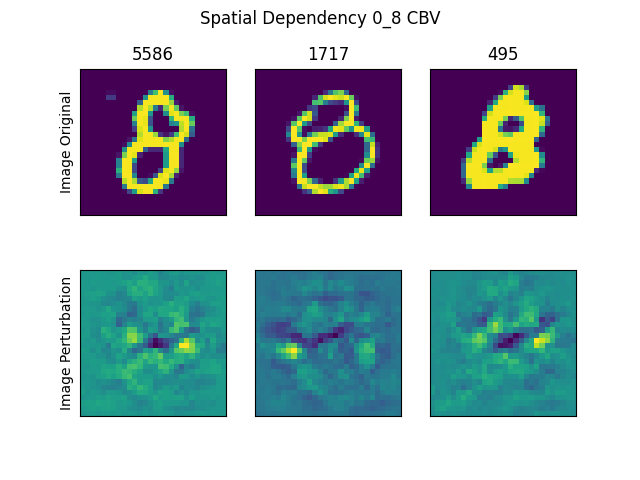}
        \caption{}
        \label{fig:spatial_dependency_cbv}
     \end{subfigure}
        \caption{Figure \ref{fig:spatial_dependency_cav} shows the spatial dependency of the concept activation vector for the $\mathsf{8}$ to $\mathsf{0}$ relationship. Figure \ref{fig:spatial_dependency_cbv} shows that spatial dependency of the concept boundary vector for the $\mathsf{8}$ to $\mathsf{0}$ relationship.}
        \label{fig:spatial_dependency}
\end{figure}

From Figure \ref{fig:spatial_dependency} we observe that both concept vectors are (negatively) influential in the centre. This is perhaps to be expected since to transform an $\mathsf{8}$ into a $\mathsf{0}$ it is sufficient to ablate the central line of the $\mathsf{8}$.

\section{The Assumptions of Concept Vectors}\label{sec:top_analysis_concept_vetors}

We now turn to verifying the validity of assumptions A1 and A2 by using the tools of topological data analysis.

\subsection{Linear Separability}

\begin{figure}[ht]
     \centering
     \begin{subfigure}[b]{0.45\textwidth}
        \centering
        \includegraphics[width=0.75\linewidth]{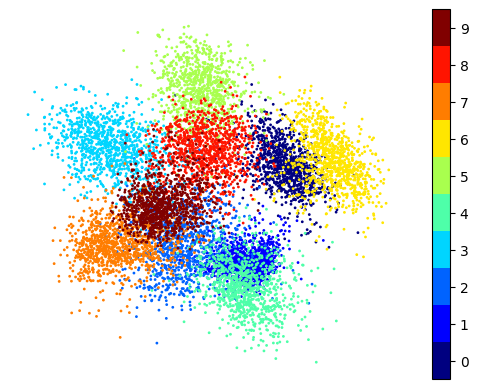}
        \caption{}
        \label{fig:svd_plot}
     \end{subfigure}
     \hfill
     \begin{subfigure}[b]{0.45\textwidth}
        \centering
        \includegraphics[width=0.75\linewidth]{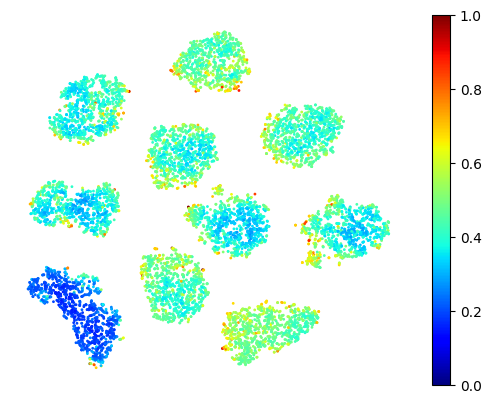}
        \caption{}
        \label{fig:tsne_plot_with_euclidicity}
    \end{subfigure}
    \caption{Figure \ref{fig:svd_plot} is a two-dimensional singular value decomposition of the latent activations of the input data on our convolutional neural network; with each point coloured according to its label. Figure \ref{fig:tsne_plot_with_euclidicity} is a t-SNE of the latent activations of the input data on our convolutional neural network; with each point coloured by its (normalised) Euclidicity value.}
    \label{fig:decomposition_plots}
\end{figure}

In most cases, the linear separability of the concept of latent representations is not an issue. Since the concepts under consideration are usually pertinent to the task performed by the machine learning model, and so since machine learning models are typically over-parameterised the latent representations of the concepts are successfully linearly distinguished by the model. Indeed, our convolutional neural network trained on MNIST successfully separates the concepts of interest (digits) as evidenced by Figure \ref{fig:svd_plot}.

The extent to which the linear separability assumption holds can be gauged by the complexity of the boundary. We obtain a proxy for this by summing the lifetimes of the zeroth homology group, $H_0$, of points along the boundary between concepts \cite{ramamurthy_topological_2018}. We then compare this to the logit influence of the concept vectors to determine the relationship between the boundary complexity and the effectiveness of the concept vectors. Specifically, for each concept we consider it as the positive concept and compute the logit influence for the concept vectors whose sources are one of the remaining concepts. We normalise the logit influence for a particular positive concept by the maximum term encountered in the summand of \eqref{eq:logit_influence} across each of the negative concepts. We then consider this against the sum of $H_0$ lifetimes for each of the concept-concept boundaries. 

\begin{figure}[ht]
    \centering
    \includegraphics[width=0.3\linewidth]{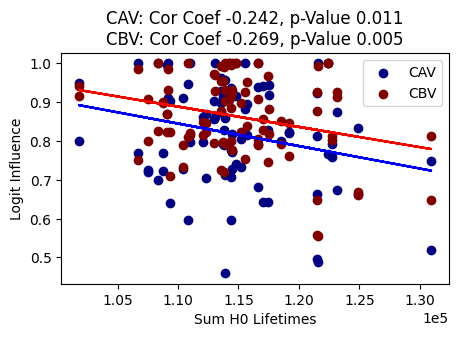}
    \caption{The (normalised) logit influence against the sum of $H_0$ lifetimes for the concept-concept boundary.}
    \label{fig:boundary_topology_logit_influence}
\end{figure}

From Figure \ref{fig:boundary_topology_logit_influence} we determine that logit influence and boundary complexity are negatively correlated. The negative correlation is significant at the $5$\% level for both concept vector formalisms, however, the negative correlation is most significant for concept boundary vectors, which is to be expected by their construction.

This shows that the boundary between concepts is influential in being able to represent the relation between concepts using concept vectors. This further motivates our concept boundary vector construction as it directly takes into account the geometry of the boundary between the concepts, whereas concept activation vectors are more agnostic to this.

\subsection{Concept Homogeneity}\label{sec:concept_homogeneity}

For a concept vector to be semantically meaningful the latent representations of a concept must exhibit a certain degree of uniformity. If the topological structure of the latent representation of a concept contains irregularities, then this suggests one of the following.
\begin{enumerate}
    \item At the stage at which the latent representations have been extracted, the model has not identified the concept as significant to its task.
    \item There exists a salient sub-concept within the concept representation that is necessary for the model to distinguish to perform well on its task.
    \item The model is utilising the concept is a non-linear manner.
\end{enumerate}

In any case, it is unlikely that a concept vector, which is interpreted as a direction in the latent space, could meaningfully capture the semantics of the concept.

We explore the homogeneity of the concept representations derived from our convolutional neural network by applying a singularity detection tool, TARDIS \cite{von_rohrscheidt_topological_2023}. TARDIS measures the \textit{Euclidicity} of a point, which captures the extent to which a given neighbourhood of that point deviates from being Euclidean. Therefore, higher Euclidicity values indicate that the point is more \textit{singular}.

From Figure \ref{fig:tsne_plot_with_euclidicity} we observe that Euclidicity within the interior of the concept clusters is fairly uniform, with boundary points being more \textit{singular}.

\begin{figure}[p]
     \centering
     \begin{subfigure}[b]{0.45\textwidth}
        \centering
        \includegraphics[width=0.6\linewidth]{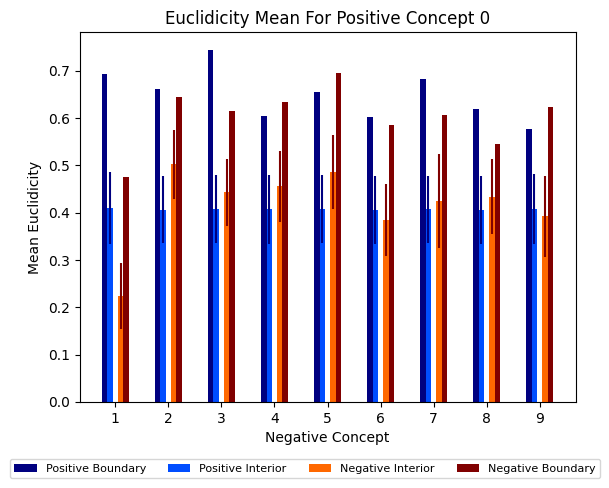}
        \caption{}
        \label{fig:mean_euclidicity_0}
     \end{subfigure}
     \hfill
     \centering
     \begin{subfigure}[b]{0.45\textwidth}
        \centering
        \includegraphics[width=0.6\linewidth]{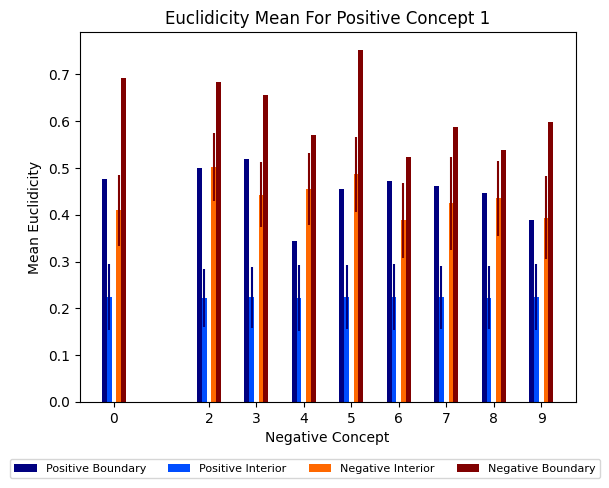}
        \caption{}
        \label{fig:mean_euclidicity_1}
     \end{subfigure}
    \hfill
     \centering
     \begin{subfigure}[b]{0.45\textwidth}
        \centering
        \includegraphics[width=0.6\linewidth]{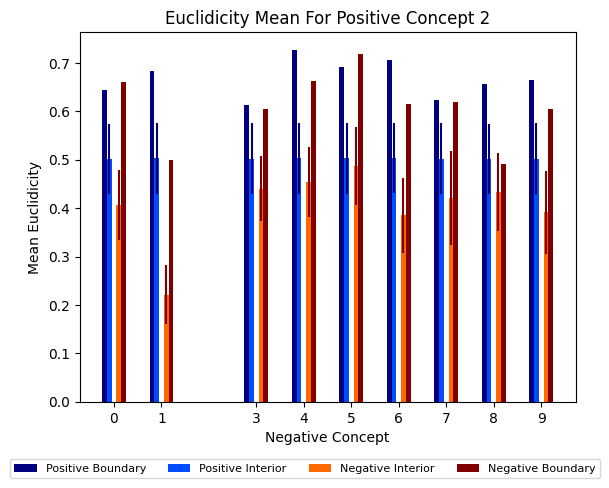}
        \caption{}
        \label{fig:mean_euclidicity_2}
     \end{subfigure}
     \hfill
     \centering
     \begin{subfigure}[b]{0.45\textwidth}
        \centering
        \includegraphics[width=0.6\linewidth]{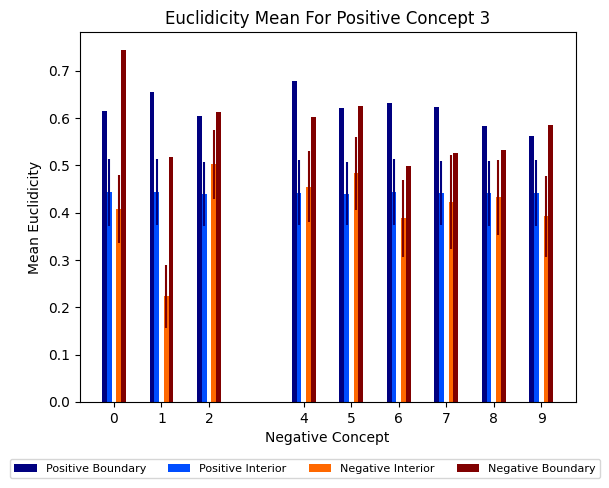}
        \caption{}
        \label{fig:mean_euclidicity_3}
     \end{subfigure}
    \hfill
     \centering
     \begin{subfigure}[b]{0.45\textwidth}
        \centering
        \includegraphics[width=0.6\linewidth]{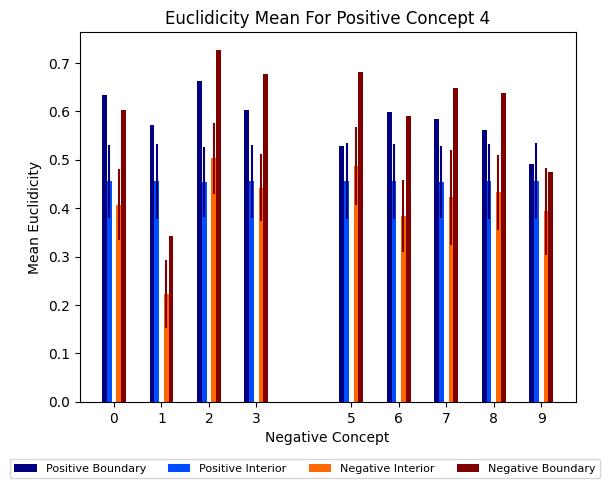}
        \caption{}
        \label{fig:mean_euclidicity_4}
     \end{subfigure}
     \hfill
     \centering
     \begin{subfigure}[b]{0.45\textwidth}
        \centering
        \includegraphics[width=0.6\linewidth]{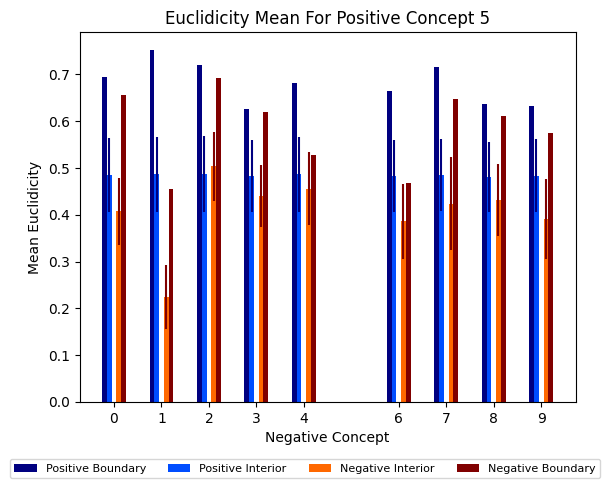}
        \caption{}
        \label{fig:mean_euclidicity_5}
     \end{subfigure}
    \hfill
     \centering
     \begin{subfigure}[b]{0.45\textwidth}
        \centering
        \includegraphics[width=0.6\linewidth]{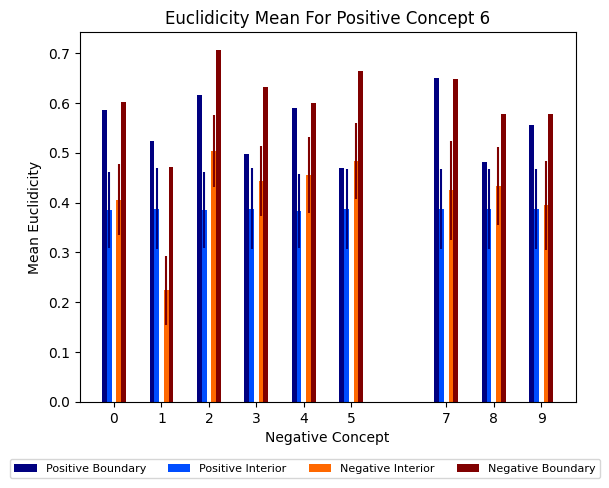}
        \caption{}
        \label{fig:mean_euclidicity_6}
     \end{subfigure}
     \hfill
     \centering
     \begin{subfigure}[b]{0.45\textwidth}
        \centering
        \includegraphics[width=0.6\linewidth]{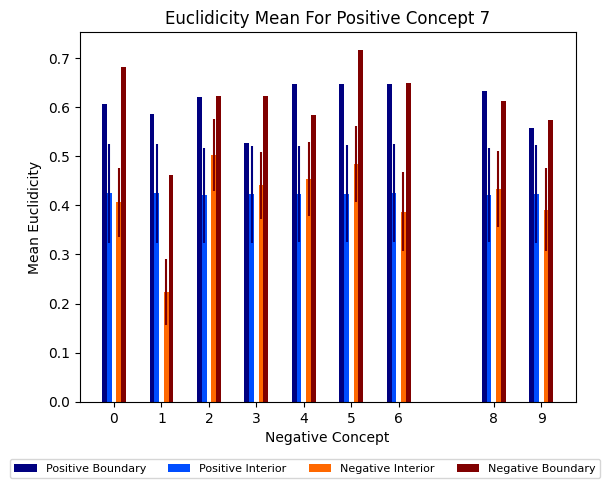}
        \caption{}
        \label{fig:mean_euclidicity_7}
     \end{subfigure}
    \hfill
     \centering
     \begin{subfigure}[b]{0.45\textwidth}
        \centering
        \includegraphics[width=0.6\linewidth]{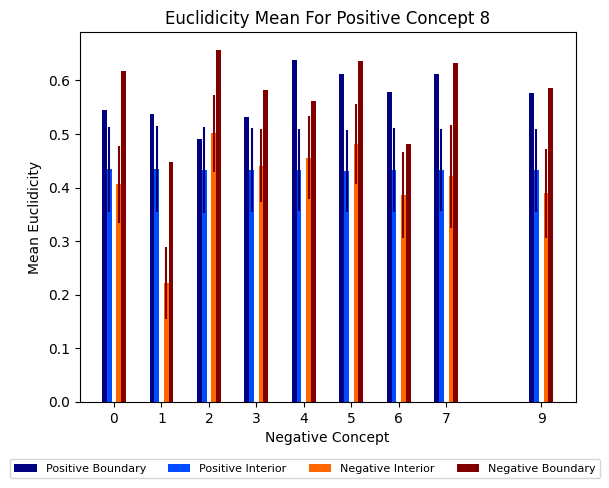}
        \caption{}
        \label{fig:mean_euclidicity_8}
     \end{subfigure}
     \hfill
     \centering
     \begin{subfigure}[b]{0.45\textwidth}
        \centering
        \includegraphics[width=0.6\linewidth]{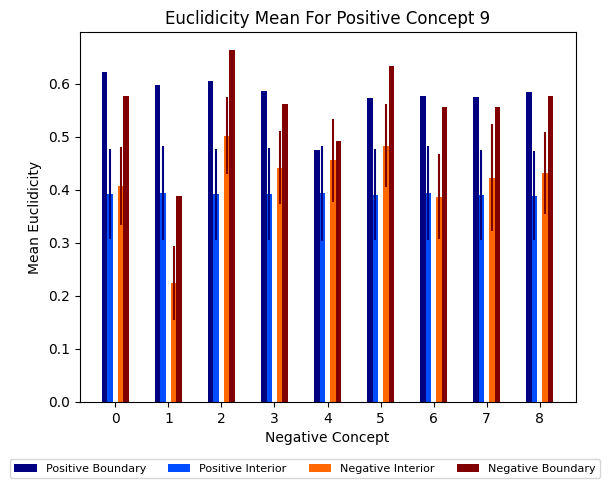}
        \caption{}
        \label{fig:mean_euclidicity_9}
     \end{subfigure}
    \hfill
    \caption{Mean Euclidicity of the interior and boundary of concept clusters. Euclidicity values have been normalised by the maximum Euclidicity across all clusters. The error bars on the interior mean bars are one standard deviation in length.}
    \label{fig:mean_euclidicity_plots}
\end{figure}

We verify that the interior of the clusters is more \textit{regular} than the boundary of the clusters by computing their respective mean Euclidicity, Figure \ref{fig:mean_euclidicity_plots}. We now explore how regularity in a concept's activations relates to its capacity to be represented by a concept vector. Specifically, we compare the logit influence of concept vectors to one of three variables.
\begin{enumerate}
    \item The difference in interior and boundary Euclidicity of the positive concept.
    \item The difference in interior Euclidicity of the positive and negative concept.
    \item The difference in boundary Euclidicity of the positive and negative concept.
\end{enumerate}

We use the same methodology as that done to obtain Figure \ref{fig:boundary_topology_logit_influence}, except we consider the logit influences against one of the three variables indicated above.

\begin{figure}[ht]
    \centering
    \includegraphics[width=0.8\linewidth]{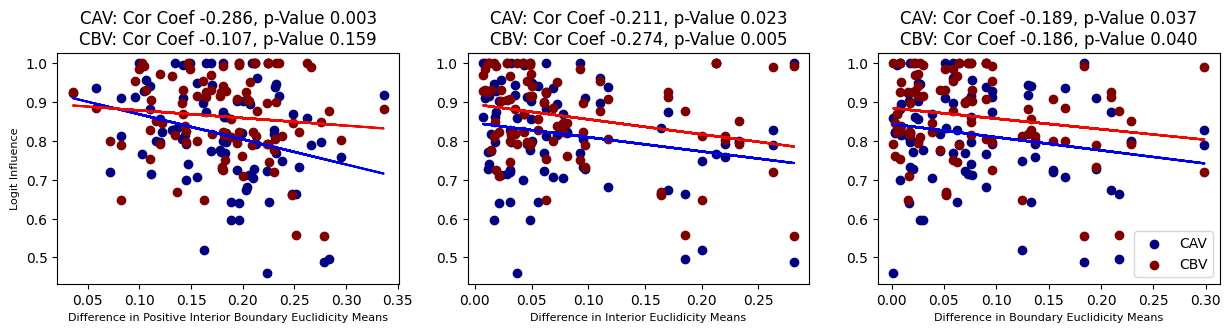}
    \caption{The (normalised) logit influence against different measures capturing the homogeneity of the concept's latent representations.}
    \label{fig:difference_in_means_avg_target_gradient_plot}
\end{figure}

We observe that in general, the logit influence of concept vectors is higher when there is homogeneity across the concept latents. In most cases, this association is statistically significant, with the most significant results being obtained when considering the difference in interior Euclidicity between the source and target activations. Suggesting that concept vectors are most effective when the positive and negative concept representations are \textit{aligned}.

\section{Conclusion}

We have introduced concept boundary vectors as a concept vector construction that is more faithful to the geometry of the boundary between concepts. Empirically, we showed that compared to concept activation vectors, concept boundary vectors are more effective at encoding the relationship between concepts and better represent the entanglement between the concepts in the latent representation of the model. Hence, we have shown that the boundary between concepts significantly influences their relationships. Our intention is that concept boundary vectors can be used as a more reliable technique for interpreting the model's representations of concepts.

Furthermore, we have taken a topological perspective on the validity of the assumptions that underline the construction of concept vectors. In particular, we have been able to relate the topological properties of the latent representations of concepts to the effectiveness of their corresponding concept vectors. We verified that concept vectors are significantly influenced by the nature of the boundary separating the concepts, and by the homogeneity of the latent representations of the concepts.

\section{Acknowledgments}

Special thanks to Professor Bastian Rieck for providing support and supervision throughout every phase of the project leading to this work.

\bibliographystyle{unsrt}  
\bibliography{references}

\section{Appendix}

\subsection{Boundary Construction Algorithm}\label{sec:boundary_construction_algorithm}

\begin{algorithm}[p]
\caption{Boundary Construction}\label{alg:boundary_construction}
\begin{algorithmic}
\Require PosConActivations, NegConActivations
\State PairsFromPositive$\gets[]$
\State CloseNegActivations$\gets[]$
\For{PosConActivation in PosConActivations}
\State ClosestNegConActivation$\gets\mathrm{GetClosestVector}$(PosConActivation,NegConActivations)
\State $\mathrm{APPEND}$(PairsFromPositive,[PosConActivation,ClosestNegConActivation])
\State $\mathrm{APPEND}$(CloseNegActivations,ClosestNegConActivation)
\EndFor
\State PairsFromNegative$\gets[]$
\State ClosePosActivations$\gets[]$
\For{NegConActivation in NegConActivations}
\State ClosestPosConActivation$\gets\mathrm{GetClosestVector}$(NegConActivation,PosConActivations)
\State $\mathrm{APPEND}$(PairsFromNegative,[ClosestPosConActivation,NegConActivation])
\State $\mathrm{APPEND}$(ClosePosActivations,ClosestPosConActivation)
\EndFor
\State Pairs$\gets[]$
\For{Pair in PairsFromPositive}
\If{Pair[0] in ClosePosActivations}
\State $\mathrm{APPEND}$(Pairs,Pair)
\EndIf
\EndFor
\For{Pair in PairsFromPositive}
\If{Pair[1] in CloseNegActivations and not(Pair in Pairs)}
\State $\mathrm{APPEND}$(Pairs,Pair)
\EndIf
\EndFor
\end{algorithmic}
\end{algorithm}


In the first phase of Algorithm \ref{alg:boundary_construction}, the activations from $\mathcal{C}_{-}$ closest to the the activations of $\mathcal{C}_{+}$ are identified. Then the activations from $\mathcal{C}_{+}$ closest to the activations of $\mathcal{C}_{-}$ are identified. The second phase then combines these identifications such that we are left with pairs $\left(a_+,a_-\right)\in A_{\mathcal{C}_+,\mathcal{C}_-}$ where either,
\begin{enumerate}
    \item $a_+$ is the closest activation in $A_{\mathcal{C}_+}$ to $a_-$ and $a_-$ is the closest activation in $A_{\mathcal{C}_-}$ to some activation, which could be $a_+$, in $A_{\mathcal{C}_+}$, or
    \item $a_-$ is the closest activation in $A_{\mathcal{C}_-}$ to $a_-$ and $a_+$ is the closest activation in $A_{\mathcal{C}_+}$ to some activation, which could be $a_-$, in $A_{\mathcal{C}_-}$.
\end{enumerate}

\begin{figure}[ht]
     \centering
     \begin{subfigure}[b]{0.45\textwidth}
        \centering
        \includegraphics[width=0.75\linewidth]{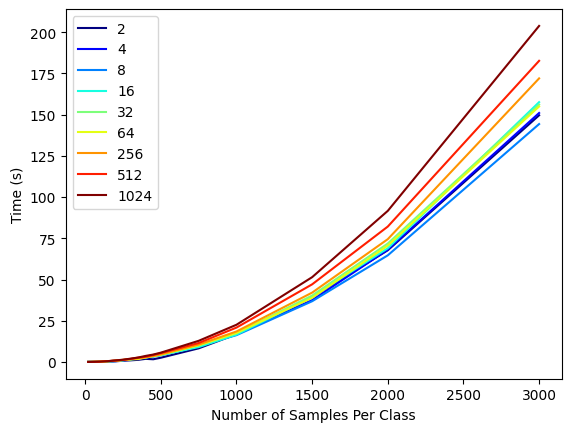}
        \caption{}
        \label{fig:algorithm_run_time_num_per_class}
     \end{subfigure}
     \hfill
     \begin{subfigure}[b]{0.45\textwidth}
        \centering
        \includegraphics[width=0.75\linewidth]{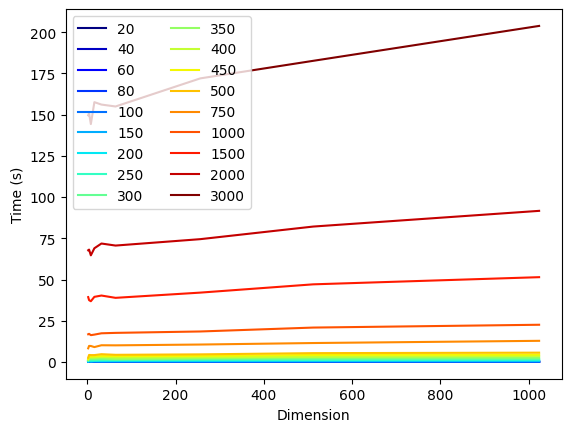}
        \caption{}
        \label{fig:algorithm_run_time_dimension}
     \end{subfigure}
        \caption{Figure \ref{fig:algorithm_run_time_num_per_class} is the run time of Algorithm \ref{alg:boundary_construction} with respect to the number of samples in the two classes, for different data dimensionalities. Figure \ref{fig:algorithm_run_time_dimension} is the run time of Algorithm \ref{alg:boundary_construction} with respect to the dimension of the data, for different numbers of samples in the two classes.}
        \label{fig:algorithm_run_time}
\end{figure}

It is clear that Algorithm \ref{alg:boundary_construction} is quadratic in the number of points in $A_{\mathcal{C}_+}$ and $A_{\mathcal{C}_-}$, however, there is also a burden imposed by the dimensionality of the latent representations. From Figure \ref{fig:algorithm_run_time_dimension} it seems as though the run time of the Algorithm \ref{alg:boundary_construction} depends linearly on the dimension of the latent representations.

\subsection{Assumption Verification Methodology}

\subsubsection{Linear Separability}

To compute the topological complexity of the boundary between concepts we use the repository of code provided by the authors of \cite{ramamurthy_topological_2018}. Specifically, we use the \verb|TopologicalData| class on the latent activations corresponding to each concept normalised by the maximum value. We labelled this data with 1 for the positive concept activations and 0 for the negative concept activations. For \verb|TopologicalData| we set the graph type equal to \verb|knn_rho|, the scale to take 100 values between 0.1 and 2, \verb|k| to equal 3, \verb|use_cy| to be true, \verb|N| to equal 20, \verb|PH_program| to be ripser, and maximum dimension to equal 1.

\subsubsection{Concept Homogeneity}

We used the tools provided in \cite{von_rohrscheidt_topological_2023} to compute the Euclidicity of our latent space activations. Specifically, we use the \verb|calculate_euclidicity| function with its default parameters except for the maximum dimension which is set to 10, the number of steps which is set to 5 and \verb|k| which is set to 40.

\subsection{Vision Transformer}\label{sec:vit}

Here we use concept vectors to explore a vision transformer model \cite{dosovitskiy_image_2021} fine-tuned on CIFAR10.\footnote{\url{https://huggingface.co/aaraki/vit-base-patch16-224-in21k-finetuned-cifar10}} Specifically, we use the concept token from each layer of the vision transformer to construct our concept vectors. The concepts we consider are the categories present in the CIFAR10 dataset. We perform this analysis to explore how concept vectors evolve across the layers of a model.

One potential limitation of concept boundary vectors exposed in this analysis is that the number of pairs identified by Algorithm \ref{alg:boundary_construction} diminishes through the layers. Therefore, the concept boundary vector may overfit to the smaller number of boundary normal vectors. However, with further exploration, we find that the concept boundary vector is still faithful to the concept-concept relationship. We think that this observation is a feature and not a bug of the concept boundary vector construction, since the number of pairs identified by Algorithm \ref{alg:boundary_construction} is indicative of the boundary geometry.

\begin{figure}[ht]
    \centering
    \includegraphics[width=0.75\linewidth]{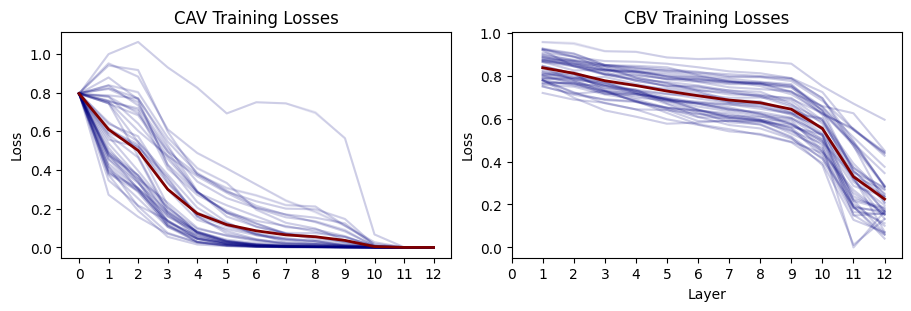}
    \caption{The training losses obtained when optimizing for the concept vectors. The blue lines are the losses for each concept-concept relationship, and the red line is the average of these losses.}
    \label{fig:training_losses}
\end{figure}

When optimizing for the concept vectors, we observe a decline in the training losses across the layers of the network, Figure \ref{fig:training_losses}. This is evidence of the fact that the relationship between concepts is becoming more linearly separable through the model.

\begin{figure}[p]
     \centering
     \begin{subfigure}[b]{0.9\textwidth}
         \centering
         \includegraphics[width=\textwidth]{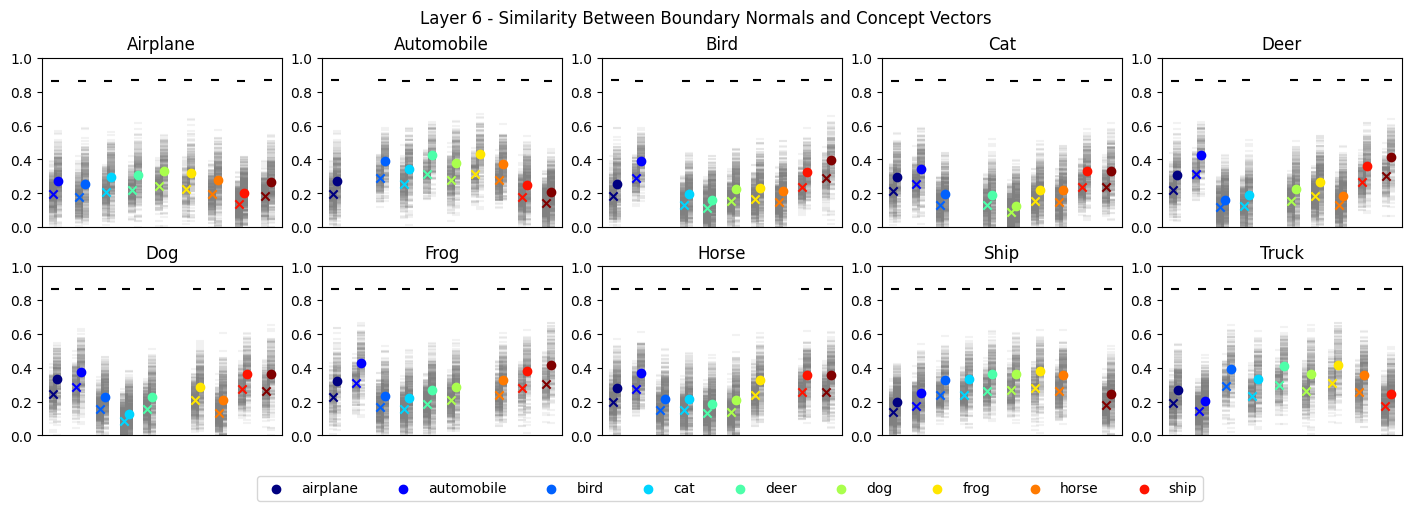}
         \caption{Similarity between boundary normal vectors and concept vectors at layer $6$ of the vision transformer model.}
         \label{fig:similarity_normals_concept_vectors_layer_6}
     \end{subfigure}
     \hfill
     \begin{subfigure}[b]{0.9\textwidth}
         \centering
         \includegraphics[width=\textwidth]{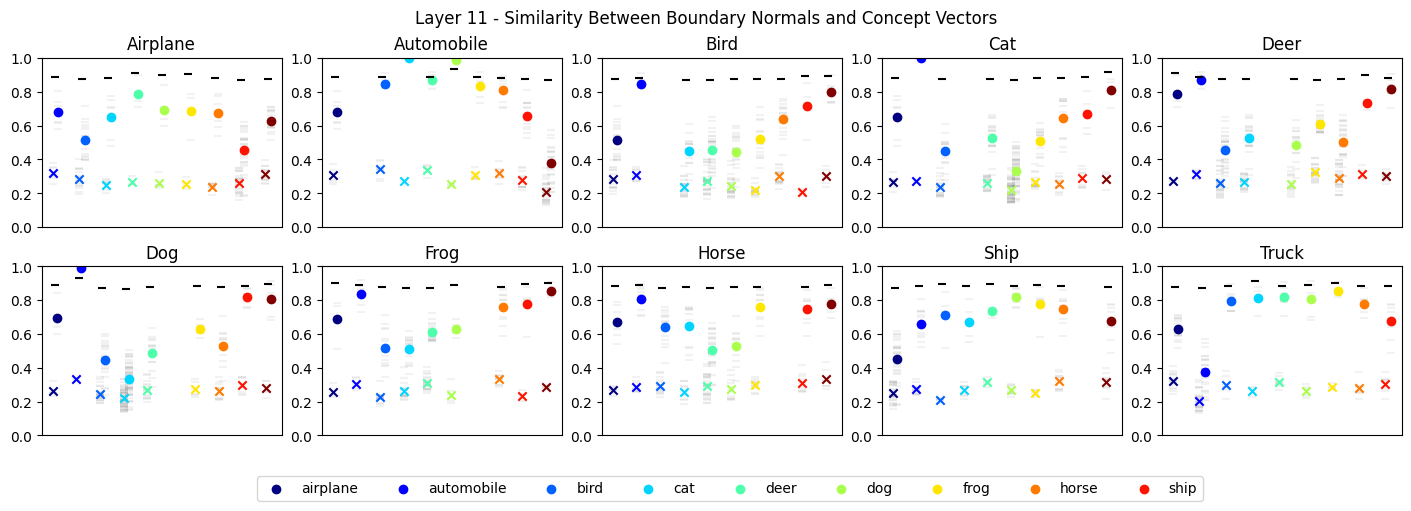}
         \caption{Similarity between boundary normal vectors and concept vectors at layer $11$ of the vision transformer model.}
         \label{fig:similarity_normals_concept_vectors_layer_11}
     \end{subfigure}
        \caption{The crosses are the average similarities to the concept activation vectors, whereas the dots are the average similarities to the concept boundary vectors. The grey lines are the individual similarities between each boundary normal vector and the concept vector. The black lines are the average similarities of a concept boundary vector optimised on random unit vectors.}
        \label{fig:boundary_normal_concept_vector_similarities}
\end{figure}

From Figure \ref{fig:boundary_normal_concept_vector_similarities} we see that concept boundary vectors are more similar to the boundary normal vectors than the concept activation vectors, as expected. In particular, the improvement in this similarity is more significant in later layers. This could either be due to there being fewer boundary normal vectors in later layers or a change in the geometry of the boundary in the later layers.

\subsubsection{Consistency}

\begin{figure}[ht]
     \centering
     \begin{subfigure}[b]{0.58\textwidth}
         \centering
         \includegraphics[width=0.75\textwidth]{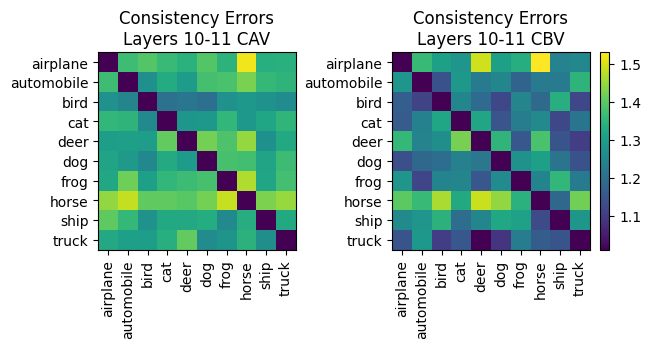}
         \caption{Consistency errors for the different concept vectors.}
         \label{fig:consistency_error_heatmaps}
     \end{subfigure}
     \hfill
     \begin{subfigure}[b]{0.38\textwidth}
         \centering
         \includegraphics[width=0.5\textwidth]{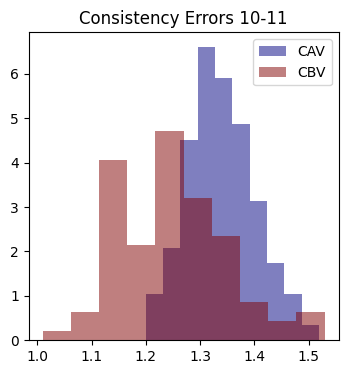}
         \caption{The distribution of the consistency errors.}
         \label{fig:consistency_errors_hist}
     \end{subfigure}
        \caption{Consistency errors between layers $10$ and $11$ of our vision transformer model.}
        \label{fig:consistency_error}
\end{figure}

A consistency error metric was introduced in \cite{nicolson_explaining_2024} to measure to what extent concept vectors across layers of a model conform with the evolution of the latent representations of the concepts through the layers. Specifically, for concepts vectors $v_{\pm}^{\ell_1}$ and $v_{\pm}^{\ell_2}$ extracted at layers $\ell_1$ and $\ell_2$ respectively, where $\ell_1<\ell_2$, the consistency error is $$\epsilon_{\text{consistency}}=\frac{1}{\left\vert S_{\mathcal{C_{+}}}\right\vert}\sum_{s\in S_{\mathcal{C}_{+}}}\left\Vert f_{\ell_1,\ell_2}\left(f_{0,\ell_1}(s)+\rho_{\ell_1}v^{\ell_1}_{\pm}\right)-\left(f_{0,\ell_2}(s)+\rho_{\ell_2}v^{\ell_1}_{\pm}\right)\right\Vert_2,$$where the notation $f_{n,m}$ represent the function of the neural network from layer $n$ to layer $m$. Here $\rho_{\ell_1}$ and $\rho_{\ell_2}$ are scaling factors that ensure the perturbed latent representations remain in distribution. We set these factors with the convention outlined in \cite{nicolson_explaining_2024}, namely $$\rho_l=\frac{\frac{1}{\left\vert S_{\mathcal{C}_{+}}\right\vert}\sum_{s\in S_{\mathcal{C}_{+}}}\left\Vert f_{0,\ell}(s)\right\Vert_2}{\left\Vert v_{\pm}^\ell\right\Vert_2}$$From Figure \ref{fig:consistency_error} we see that concept boundary vectors are typically more consistent between the layers.

\begin{figure}[p]
     \centering
     \begin{subfigure}[b]{0.45\textwidth}
         \centering
         \includegraphics[width=\textwidth]{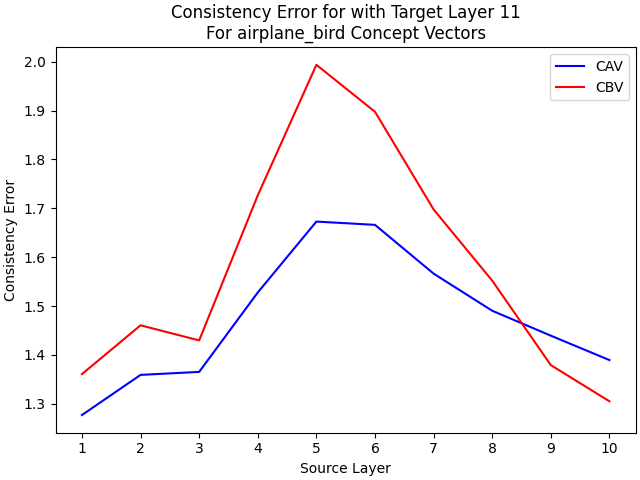}
         \caption{The consistency errors of the aeroplane-bird concept vectors through the layers of the vision transformer model.}
         \label{fig:consistency_errors_airplane_bird}
     \end{subfigure}
     \hfill
     \begin{subfigure}[b]{0.45\textwidth}
         \centering
         \includegraphics[width=\textwidth]{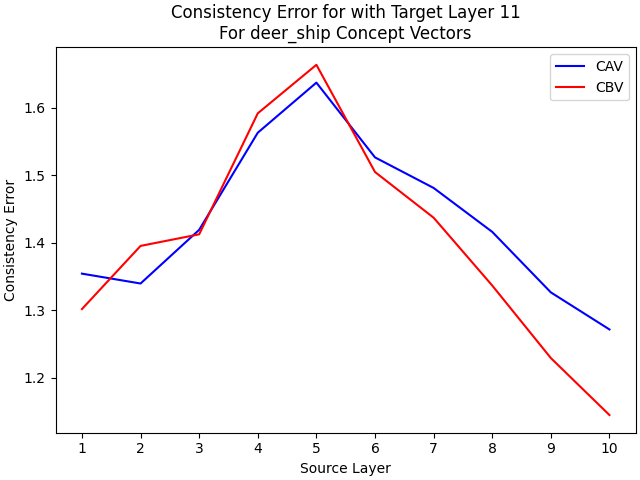}
         \caption{The consistency errors of the deer-ship concept vectors through the layers of the vision transformer model.}
         \label{fig:consistency_errors_deer_ship}
     \end{subfigure}
     \hfill
     \begin{subfigure}[b]{0.45\textwidth}
         \centering
         \includegraphics[width=\textwidth]{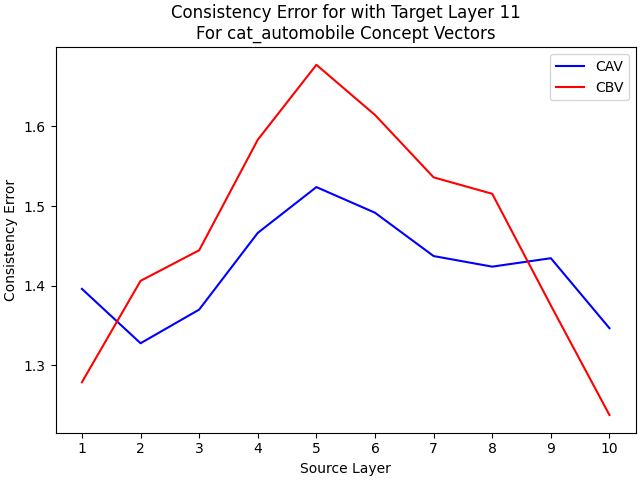}
         \caption{The consistency errors of the cat-automobile concept vectors through the layers of the vision transformer model.}
         \label{fig:consistency_errors_cat_automobile}
     \end{subfigure}
     \hfill
     \begin{subfigure}[b]{0.45\textwidth}
         \centering
         \includegraphics[width=\textwidth]{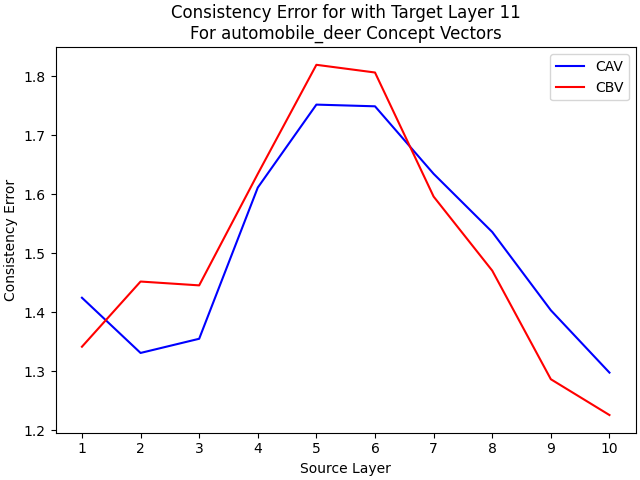}
         \caption{The consistency errors of the automobile-deer concept vectors through the layers of the vision transformer model.}
         \label{fig:consistency_errors_automobile_deer}
     \end{subfigure}
        \caption{}
        \label{fig:consistency_error_through_layers}
\end{figure}

Interestingly, by observing the consistency errors of a concept vector across the layers of the model, we see that the error peaks consistently at layers 5 and 6 before tapering towards its minimum value, Figure \ref{fig:consistency_error_through_layers}. A possible explanation is that in the earlier layers, the model organises the latent representation according to high-level features which remain consistent. Then in the middle layers, the model is trying to disentangle the more intricate details leading to a spike in the consistency error. The model then refines these details in the later layers which causes the consistency error to decrease to its minimum value. In most cases, this pattern is most pronounced with the consistency errors associated with the concept boundary vector. We test this hypothesis by training classifiers on the representations of each layer of the vision transformer.

\begin{figure}[ht]
    \centering
    \includegraphics[width=0.6\linewidth]{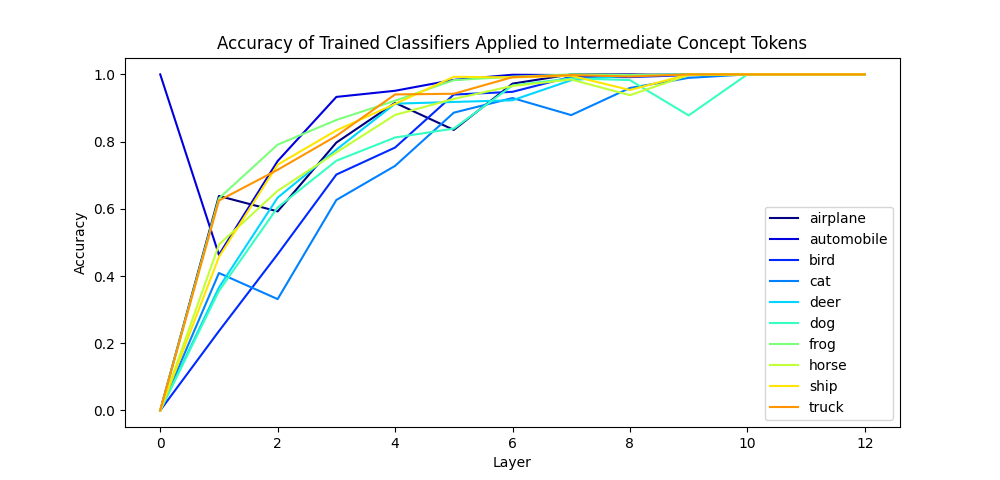}
    \caption{The accuracy of a trained classifier applied to the concept activations at intermediate layers through the vision transformer model.}
    \label{fig:trained_classifier_intermediate_accuracies}
\end{figure}

From Figure \ref{fig:trained_classifier_intermediate_accuracies} we observe that by layer 6 the latent representations of the concept are delineated to the extent that a classifier can distinguish the concepts with relatively high accuracy. By just observing the accuracies it is unclear what the model is doing from layer 6 to its output layer. However, from our consistency error analysis, it is clear that that model is manipulating the latent representations to further delineate the intricate features of each concept.

\subsubsection{Semantic Representation}

\begin{figure}[p]
     \centering
     \begin{subfigure}[b]{0.45\textwidth}
         \centering
         \includegraphics[width=\textwidth]{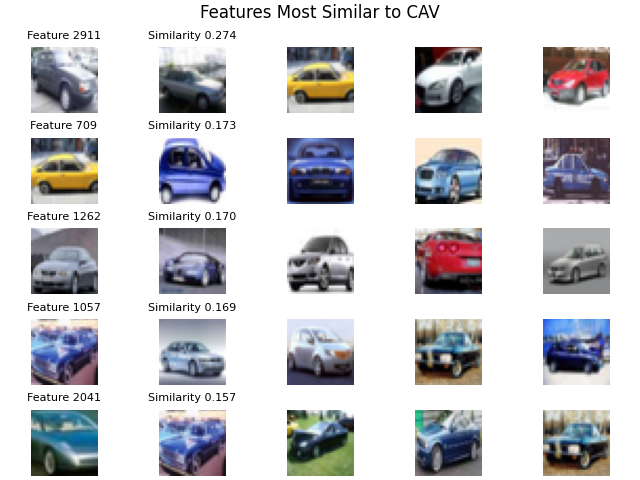}
         \caption{Images that fire the most similar features to the automobile-horse concept activation vector.}
         \label{fig:most_similar_features_cav}
     \end{subfigure}
     \hfill
     \begin{subfigure}[b]{0.45\textwidth}
         \centering
         \includegraphics[width=\textwidth]{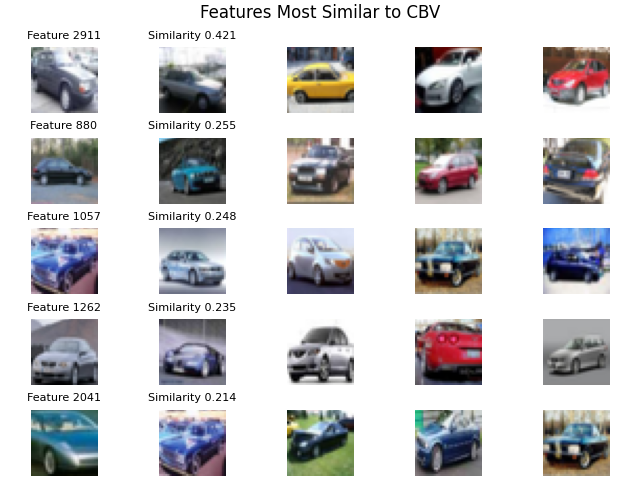}
         \caption{Images that fire the most similar features to the automobile-horse concept boundary vector.}
         \label{fig:most_similar_features_cbv}
     \end{subfigure}
     \hfill
     \begin{subfigure}[b]{0.45\textwidth}
         \centering
         \includegraphics[width=\textwidth]{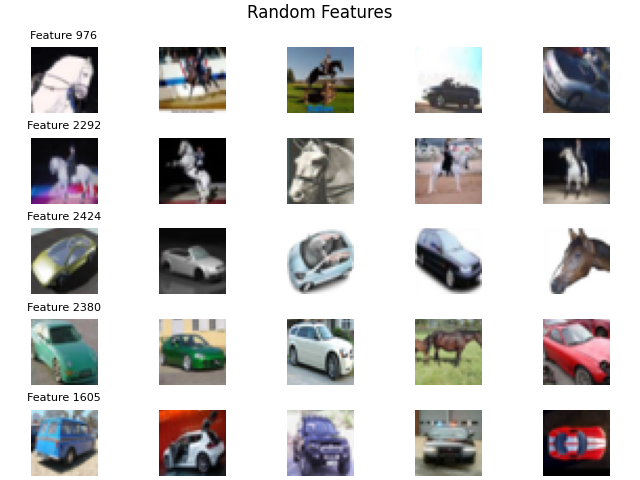}
         \caption{Images that fire for random features.}
         \label{fig:random_features}
     \end{subfigure}
        \caption{Identifying the images that trigger the features extracted by the sparse autoencoder trained on the latent representations of the automobile and horse concept images at layer $11$ of the model.}
        \label{fig:layer11_features_similar_to_concept_vectors}
\end{figure}

To extract features from the latent space of our vision transformer we use sparse autoencoders. These have been shown, in the context of large language models \cite{trenton_bricken_towards_2023}, to extract interpretable features from the latent representations in an unsupervised manner. The sparse autoencoder we use has the architecture of a tied weights autoencoder \cite{cunningham_sparse_2023} and is trained using mean square error with a sparsity regularisation term. Through this process, we construct a dictionary of features for the latent representations. We compute the cosine similarity of these features with our concept vectors and identify the images in the training set that cause these features to fire most strongly. Some results of this analysis applied to layer $11$ of the vision transformer can be found in Figure \ref{fig:layer11_features_similar_to_concept_vectors}.

\begin{figure}[p]
     \centering
     \begin{subfigure}[b]{0.45\textwidth}
         \centering
         \includegraphics[width=\textwidth]{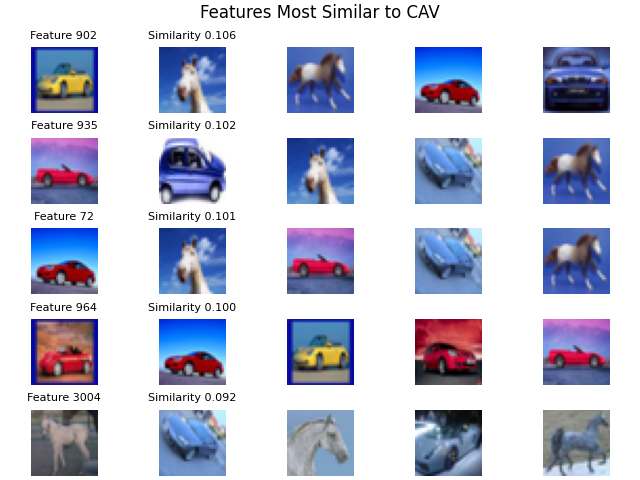}
         \caption{Images that fire the most similar features extracted at layer $1$ to the automobile-horse concept activation vector obtained at layer $1$.}
         \label{fig:most_similar_features_cav_layer1}
     \end{subfigure}
     \hfill
     \begin{subfigure}[b]{0.45\textwidth}
         \centering
         \includegraphics[width=\textwidth]{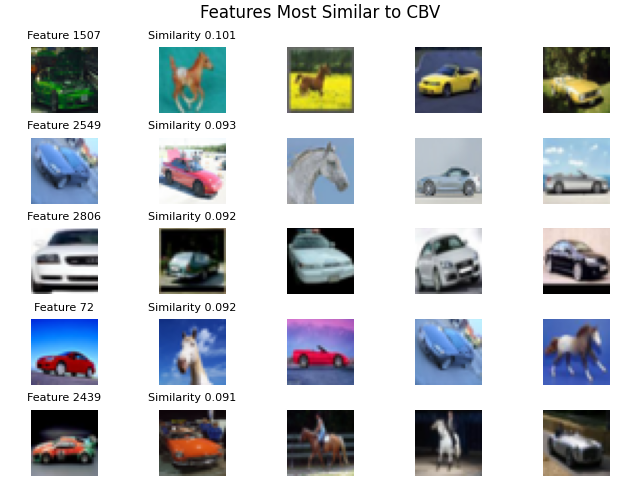}
         \caption{Images that fire the most similar features extracted at layer $1$ to the automobile-horse concept boundary vector obtained at layer $1$.}
         \label{fig:most_similar_features_cbv_layer1}
     \end{subfigure}
     \hfill
     \begin{subfigure}[b]{0.45\textwidth}
         \centering
         \includegraphics[width=\textwidth]{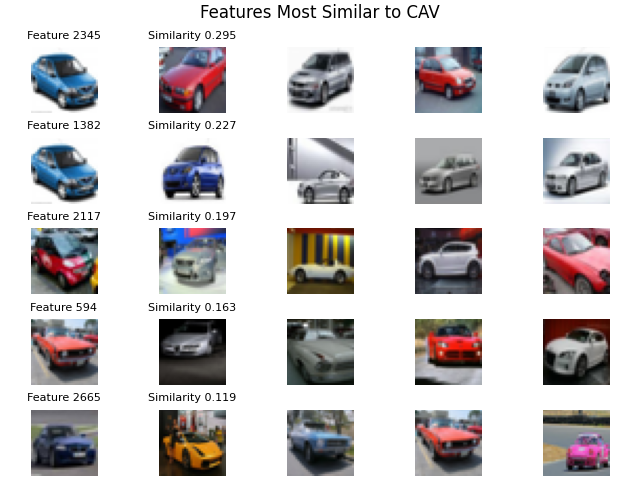}
         \caption{Images that fire the most similar features extracted at layer $6$ to the automobile-horse concept activation vector obtained at layer $6$.}
         \label{fig:most_similar_features_cav_layer6}
     \end{subfigure}
     \hfill
     \begin{subfigure}[b]{0.45\textwidth}
         \centering
         \includegraphics[width=\textwidth]{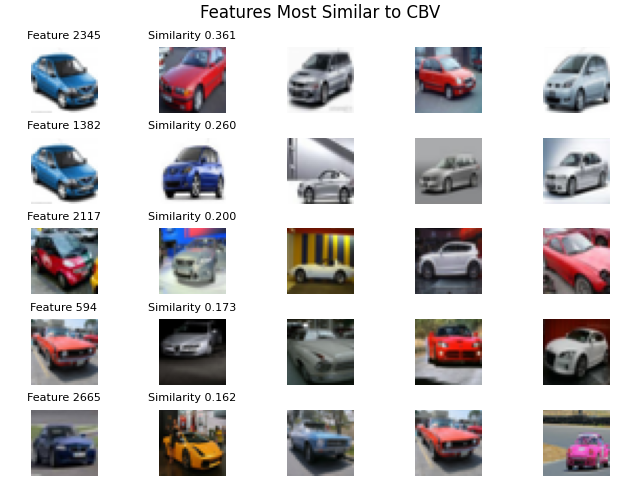}
         \caption{Images that fire the most similar features extracted at layer $6$ to the automobile-horse concept boundary vector obtained at layer $6$.}
         \label{fig:most_similar_features_cbv_layer6}
     \end{subfigure}
        \caption{Identifying the images that trigger features extracted by the sparse autoencoder trained on the latent representations of the automobile and horse concept images at layers $1$ and $6$ of the model.}
        \label{fig:layer1/6_features_similar_to_concept_vectors}
\end{figure}

From Figure \ref{fig:layer11_features_similar_to_concept_vectors} we see that both concept activation vectors and concept boundary vectors encode some semantic meaning of their target concept, as they are able to identify features of the target concept not present when we randomly sample features. In particular, we note that the features are more similar to the concept boundary vector. We can repeat this analysis at different layers of the model, doing so we obtain the results of Figure \ref{fig:layer1/6_features_similar_to_concept_vectors}.

From Figure \ref{fig:layer1/6_features_similar_to_concept_vectors} we observe that the similarity between the concept vectors and the features increases through the layers. In layer $6$ we again observe that the concept boundary vector is more similar to the features than the concept activation vectors. However, in all of these experiments, the cosine similarity between the concept vectors and the features is relatively small. Perhaps suggesting that the Euclidean inner product is not the optimal metric to capture a notion of similarity. This idea is explored further in \cite{park_linear_2024} where for a collection of concepts a transformation is identified such that the Euclidean inner product in this transformed space effectively represents a notion of similarity.

\end{document}